# Improving Drug Identification in Overdose Death Surveillance using Large Language Models

*Authors:*

*Arthur J. Funnell, BSc[a], Panayiotis Petousis, PhD[a], Fabrice Harel-Canada, PhD[b], Ruby Romero, BA[c], Alex A. T. Bui, PhD[a], Adam Koncsol, MPA[d], Hritika Chaturvedi[c], Chelsea Shover, PhD[d], David Goodman-Meza, PhD[e]*

*Affiliations:*

[a]Medical & Imaging Informatics Group, Department of Radiological Sciences, David Geffen School of Medicine, University of California Los Angeles, Los Angeles, CA

[b]Computer Science Department, University of California, Los Angeles, 404 Westwood Plaza Suite 277, Los Angeles, 90095, CA, USA

[c]Computational and Systems Biology, University of California, Los Angeles, 102 Hershey Hall, Box 951600, Los Angeles, 90095, CA, USA

[d]Division of General Internal Medicine and Health Services Research, University of California, Los Angeles, 1100 Glendon Ave STE 850, Los Angeles, 90024, CA, USA

[e]Kirby Institute, University of New South Wales, Wallace Wurth Building (C27), Cnr High St & Botany St, UNSW, Sydney, 2052, NSW, Australia

## Conflict of Interest Statement

The authors have no conflicts of interest to declare.

## Funding Sources

CLS was supported by a grant from the National Institute on Drug Abuse (K01-DA050771, PI: Shover). All authors were supported National Institute on Drug Abuse (R01-DA57630, PIs: Shover, Goodman-Meza). The funders had no role in the design, conduct, or decision to publish this manuscript.

## Acknowledgments

The authors would like to acknowledge Nanyun (Violet) Peng for providing the resources to enable F. Harel-Canada to finetune and run the LLMs.

## Author Contributions

A. Funnell conceptualized the study, performed all analysis excluding work with LLMs (Qwen 3.1 and Llama 3.2). A. Funnell worked on the investigation, methodology, project administration, resources, software, validation, visualization. A. Funnell wrote, reviewed, and edited the manuscript.

P. Petousis conceptualized the study and assisted with analysis. P. Petousis worked on the investigation, methodology, resources, software, and project administration. P. Petousis wrote, reviewed, and edited the manuscript.

F. Harel-Canada performed analysis with LLMs (Qwen 3.1 and Llama 3.2). F. Harel-Canada worked on the methodology, resources, and software. F. Harel-Canada wrote, reviewed, and edited the manuscript

R. Romero worked on data curation, and project administration. R. Romero wrote, reviewed and edited the manuscript.

A. A. T. Bui conceptualized the study. A. A. T. Bui, wrote, reviewed, and edited the manuscript.

H. Chaturvedi worked on Data curation. H. Chaturvedi, reviewed, and edited the manuscript.

A. Koncsol worked on data curation. A. Koncsol reviewed and edited the manuscript.

C. Shover conceptualized the study. C. Shover oversaw curation of both datasets. C. Shover worked on funding acquisition and project administration. C. Shover wrote, reviewed and edited the manuscript.

D. Goodman-Meza conceptualized the study. D. Goodman-Meza reviewed analysis. D. Goodman-Meza offered supervision. D. Goodman-Meza worked on project administration and funding acquisition. D. Goodman-Meza wrote, reviewed and edited the manuscript.

## Data Sharing Statement

The data collected for this study are publicly available. All code used to train and recreate the experiments seen in the paper are available on Github.

Arthur Funnell's Github:

https://github.com/elemets

Fabrice Harel-Canada's Github:

https://github.com/fabriceyhc

All code and data for this paper:

https://github.com/ROSLA-UCLA/ROSLA-HEAL-Data/tree/main

**Email Address:** Arthur J. Funnell: afunnell@mednet.ucla.edu

**Abstract**

The rising rate of drug-related deaths in the United States, largely driven by fentanyl, requires timely and accurate surveillance. However, critical overdose data are often buried in free-text coroner reports, leading to delays and information loss when coded into ICD (International Classification of Disease)-10 classifications. Natural language processing (NLP) models may automate and enhance overdose surveillance, but prior applications have been limited.

A dataset of 35,433 death records from multiple U.S. jurisdictions in 2020 was used for model training and internal testing. External validation was conducted using a novel separate dataset of 3,335 records from 2023-2024. Multiple NLP approaches were evaluated for classifying specific drug involvement from unstructured death certificate text. These included traditional single- and multi-label classifiers, as well as fine-tuned encoder-only language models such as Bidirectional Encoder Representations from Transformers (BERT) and BioClinicalBERT, and contemporary decoder-only large language models such as Qwen 3 and Llama 3. Model performance was assessed using macro-averaged F1 scores, and 95% confidence intervals were calculated to quantify uncertainty.

Fine-tuned BioClinicalBERT models achieved near-perfect performance, with macro F1 scores ≥0.998 on the internal test set. External validation confirmed robustness (macro F1=0.966), outperforming conventional machine learning, general-domain BERT models, and various decoder-only large language models.

NLP models, particularly fine-tuned clinical variants like BioClinicalBERT, offer a highly accurate and scalable solution for overdose death classification from free-text reports. These methods can significantly accelerate surveillance workflows, overcoming the limitations of manual ICD-10 coding and supporting near real-time detection of emerging substance use trends.

**Highlights**

- NLP models accurately classify drugs in overdose death certificate text.
- Multi-label classification methods reflect real-world polysubstance use in overdose deaths.
- LLMs offer fast scalable alternatives to manual ICD-10 coding, supporting forensic toxicology.
- BioclinicalBERT achieved F1 scores of 0.999 (internal test set) and 0.965 (external test set).

**Introduction**

In the United States, the rate of drug-related deaths increased from 8.9 deaths per 100,000 in 2003 to 32.6 in 2022 (1), driven primarily by the rise of fentanyl (2). Data supporting these statistics is often found only in unstructured free-text reports written by local medical examiners and coroners (3). These reports are shared with local jurisdictions, where they are later manually coded into ICD-10 classifications. This labor-intensive process delays the reporting of critical data, slowing resource allocation (4). Moreover, the ICD-10 coding system can obscure essential details. For instance, both buprenorphine (used to treat opioid use disorder) and fentanyl (a synthetic opioid) share the same ICD-10 code (4,5). Moving beyond reliance on ICD-10 codes may reveal critical insights into population-level changes in drug-related deaths (6–8).

Natural language processing (NLP) is a field of artificial intelligence that enables computers to understand, interpret, generate, and manipulate human language in a meaningful way. While NLP has been applied in myriad medical contexts (9), research on its use to process free-text data in death certificates and death investigation reports remains limited (3,4). Recently, decoder only large language models (LLMs), such as GPT-4o and LLAMA 3, have achieved state-of-the-art performance on many text related tasks, such as document classification (10,11). However, the high compute cost, fiscal cost and complexity of these larger and newer LLMs render them less feasible for implementation in settings where specialized computing personnel and resources are limited (e.g., many medical examiner and coroner offices) (12). Encoder only models such as Bidirectional Encoder Representations from Transformers (BERT) models may be more suitable for these settings, considering their high accuracy, lower complexity, and lower cost. Currently, a significant gap exists in the literature regarding the utilization of BERT models to identify and categorize free-text data related to drugs. BERT models are a type of language model that processes text by analyzing the context of each word in both directions - before and after - to better ascertain semantic meaning (13). This allows BERT to more accurately extract relevant information from unstructured text, such as drug names and causes of death, making it particularly suitable for classifying death records in relation to drug involvement.

Some previous research has examined BERT models for categorizing death data, with varying success. The specific application of categorizing drug involvement has also been explored using other NLP techniques. Parker et al. (14) demonstrated the potential of compact language models like distilBERT, achieving an F1 score, a balance of sensitivity and positive predictive value, of 0.6 on violent death reports with only 1,500 cases. But this work focused on violent deaths rather than drug-related deaths, limiting its direct applicability to the task of classifying specific drug involvement. Ward et al. achieved an F1 score of 0.97 for identifying fatal drug overdoses in test data; however, this performance reflects only a distinction between overdose and non-overdose cases and does not attempt to classify drug type (3). Goodman et al. successfully classified drug types in text data using concept unique identifiers (CUIs), with F1 scores exceeding 0.95 for most substances (4). Despite this, the reliance on rule-based methods and exact matches for CUIs reduced accuracy in cases of spelling errors or text variations, particularly for benzodiazepines and alcohol, with F1 scores of 0.840 and 0.854, respectively (4). To our knowledge, the performance of BERT models in categorizing specific drug involvement has not been described in the literature.

In this study, we improve on Goodman et al.'s earlier work on overdose death classification. The objectives of this study were threefold: 1) evaluate the effectiveness of BERT embedding models to classify overdose deaths by reproducing the methodology of Goodman et al.; 2) extend the analysis from single-label to multi-label classification to better reflect real-world polysubstance abuse; and 3) conduct NLP model comparisons and external validation of the best-performing model using an independent test dataset.

## Methods

### Data

Death certificate data was obtained from ten different regions across the United States via publicly accessible sources or by directly requesting it from coroners and medical examiners. The areas where data was obtained included Cook County, Illinois; Denton County, Texas; Jefferson County, Alabama; Johnson County, Texas; Los Angeles County, California; Milwaukee

County, Wisconsin; Parker County, Texas; San Diego County, California; Tarrant County, Texas; and Connecticut. All records provided were included in our analysis. Information was compiled into a database containing the following variables: case number, county, age, gender, race, date of death, manner of death, primary cause, and secondary cause of death. The data spanned a one-year period, from January 1, 2020, to December 31, 2020.

A second external validation dataset was used to verify the results. This dataset consisted of death certificates from 3,335 decedents from ten US counties: Cook County, Illinois; Los Angeles County, California; Middlesex, New Jersey; Windham, New Hampshire; Tollham Connecticut; New Haven, Connecticut; Hartford, Connecticut; Fairfield, New London, Connecticut. The dataset spanned from the 1st of October 2023 to the 31$^{st}$ of April 2024.

### Reference Standard

Four authors (C.L.S., R.R., A.K., H.C.) manually classified text based on whether a substance was present in each case, based on the accompanying unstructured text from the coroner. A given case can involve multiple substances. As such, we labeled each case using a binary classification system (0 = No, 1 = Yes) to indicate whether a specific drug contributed to the decedent's death. Annotators only included drug-related deaths resulting from acute causes (e.g., toxicity), excluding those due to chronic conditions. C.LS. reviewed the annotators' first 20 annotations before they continued with their full set of assignments to check the instructions were being followed consistently and the task was properly understood.

### Modeling

The problem was approached using four methodologies (see Table 1): 1) a single-label classic machine learning approach; 2) a multi-label classic machine learning approach; 3) fine-tuning of a Transformer model (BERT) (13) multi-label approach; and 4) fine-tuning of large language models (LLMs) (15) for a multi-label approach. In the case of multi- and single-label classic machine learning models, several model architectures and embedding methods were tested. In both the single- and multi-label approaches, the input text was first preprocessed by lowercasing and removing stop words (4).

**Table 1: Different methodologies evaluated, the data splits employed, and the different embedding types and model architectures used for training.**

| Approach | Data Split (Train/Val/Test) | Embeddings | Models / Architectures |
|---|---|---|---|
| Single-label classification | Stratified 80/20 | BioclinicalBERT GloVe CUIs | Logistic Regression XGBoost Random Forest Support Vector Machine |
| Multi-label classification | 60/20/20 | BioclinicalBERT | XGBoost RandomForest |
| Finetuned encoder-only BERT models | 60/20/20 | Pretrained tokenizers from huggingface: BERT BioclinicalBERT | Fine-tuned encoder only models: BERT BioclinicalBERT |
| Finetuned decoder-only LLMs | 60/20/20 | Pretrained tokenizers from huggingface: Qwen 3 Llama 3 | Fine-tuned decoder only models: Qwen 3 Llama 3 |
| Abbreviations: BERT, Bidirectional Encoder Representations from Transformers; CUI, concept unique identifier; GloVe, Global Word Embedding Vectors; XGBoost, Extreme Gradient Boosting; LLM, large language model; ||||

**Single label**

Goodman et al.'s (4) prior work was reproduced with all experiments' model performance within the published confidence intervals. To generate text embeddings (used as input features) Global Vector for Word Representation (GloVe) - 6B 400k Wikipedia Gigaword - and Concept unique identifiers (CUIs) were employed. Specifically, GloVe uses a set of word embeddings generated from public domain content. We used the 100-dimensional vectors as embeddings derived from 6B tokens 400K vocab Wikipedia and Gigaword version (16). CUIs are alphanumeric identifiers assigned by the Unified Medical Language System (17). Text was converted to CUIs and then converted to text embeddings via the Cui2Vec package (18). The CUI2Vec embeddings were filtered to include only the semantic class of "organic chemical." We also tested an extra embedding method using the same experimental design via BioclinicalBERT. BioclinicalBERT involves word vector embeddings fine-tuned on the MIMIC III dataset of records from intensive care unit (ICU) patients, and thus may contain medical concepts related to overdose (19).

For each embedding method, we tested four model architectures: logistic regression, eXtreme Gradient Boosting (XGBoost), random forest, and support vector classifier (SVC). Binary classifiers were trained per drug. Data was split in a stratified manner regarding the drug category of interest: 80% training and validation, 20% testing. The models were trained using a 10-fold cross-validation. Hyperparameters were tuned and models trained using the Model Tuner package, which streamlined the training and evaluation of these models (20). We chose a wide hyperparameter space for each different model and performed a grid search to find the best-performing combination of hyperparameters for each model. We balanced the class weights to account for class imbalance. The area under the receiver operating characteristic curve (AUROC) was used to determine the best model. The model with the combination of hyperparameters that scored the highest AUROC on the validation set was then assessed on the test set. To assess the uncertainty of our estimates, we applied bootstrapping with 1,000 resampling iterations. When looking at the test set, we evaluated the macro-averaged F1 score, AUROC, and average precision (AP). The macro-averaged F1 score was chosen as it provides a better representation of model performance when there is class imbalance (21).

### Multi-label

The data were split into train, validation, and test sets. 60% training, 20% validation, and 20% test. Stratification was not feasible in the multi-label classification task due to numerous combinations of labels (see Figure 1), where each data point could belong to multiple substances (e.g., fentanyl, heroin, and cocaine), leading to insufficient instances for each unique label combination to perform stratified splitting. Text data were embedded using BioClinicalBERT. Both XGBoost and random forest algorithms were trained as multi-label classifiers, as these have inherent multi-label support. Various scoring functions were evaluated and optimized, with primary emphasis on average precision, the AUROC, and Hamming loss (used as it works well for imbalanced datasets). A similar grid search hyperparameter tuning methodology was applied using the Model Tuner library (20).

### Fine-tuning BERT models

Both the BERT and BioclinicalBERT models were fine-tuned using the same dataset split as used when assessing the multi-label models. The pretrained models were obtained from Hugging Face (22). Fine-tuning for the classification task involves adding a classification head and retraining the unfrozen model weights, which was performed utilizing the Transformers library (23). A batch size of 32 was chosen, a weight decay of 0.01, a learning rate of 2e-5, and the F1 metric was used to evaluate the model. Model performance was evaluated using macro-averaged F1, accuracy, and AUROC. Notably, accuracy reflects instances where all drug labels were predicted correctly - a high accuracy indicates an accurate prediction in the co-occurrences of drug intake in overdoses.

### Other Large Language Models (LLMs)

Llama 3 and Qwen 3 models were compared, ranging from 0.6B parameter models to 8B parameter models. The full set of LLMs tested can be found in Supplemental Table 3. This included evaluating the performance of each model with/without a differing number of few-shot prompt examples as well as fine-tuning the models on the training set. The models were trained via supervised fine-tuning (SFT) until the training loss fell below 0.005. For most models this occurred with just 2,000 examples from the train set. For the few-shot evaluation, we used 0, 3, 5, 10 examples from the validation set within the prompt. Further details of the methodology can be found in Harel-Canada et al.'s work on substance use detection. (15)

### Evaluating models

Model evaluation involved comparing each model against the same performance metrics on the same test set. Single-label models were combined to perform as a multi-label model, using them to predict a multi-label outcome by loading all the models for each drug class and then predicting each data point. We used this output to generate accuracy, AUROC, Hamming loss, and macro-averaged F1 scores. This enabled a direct comparison of the performance of the single-label solution against the multi-label, fine-tuned BERT, and LLMs. An external dataset was used to validate the results using the best models found. Error analysis was performed to identify difficult to classify cases and areas where the models underperformed (Supplemental Table 4, Supplemental Table 5).

## Results

### Data breakdown

The initial data set included 35,698 cases. We excluded 265 cases due to missing textual data, resulting in a final dataset of 35,433 cases. The jurisdictions that provided the most cases were Cook County (45%), Los Angeles County (32%), and San Diego County (8%). The median number of characters per text for each case was 59 (range, 3 to 331). The median number of words per text was 7 (range, 1 to 38). The number of substances or groups of substances classified were zero (no substances involved in cause of death) in 26,695 cases (75%); one in 2,635 cases (7%); two in 1,401 cases (4%); three in 2,218 cases (6%); four in 1,364 cases (4%); five in 659 cases (2%); six in 301 cases (1%); and seven in 113 cases (<1%) (4). Substances with a count below a cutoff of 1,000 (e.g., 3,4-methylenedioxymethamphetamine, 3,4-methylenedioxyamphetamine, amphetamine, antipsychotics, antidepressants, anticonvulsants, antihistamines, muscle relaxants, barbiturates, and hallucinogens) were grouped as others. The substances identified to be related to death include any opioids (5,739 [16%]), fentanyl (4,758 [13%]), alcohol (2,866 [8%]), cocaine (2,247 [6%]), methamphetamine (1,876 [5%]), heroin (1,613 [5%]), prescription opioids (1,197 [3%]), and any benzodiazepine (1,076 [3%]) (4).

The frequency of the top 20 co-occurrences of substances is displayed in Figure 1. In total, there were 4,636 (13%) examples where the overdose was caused by more than one substance. This was more than half of the total number of overdose deaths in the data (8,738). There was a total of 180 unique combinations of drug overdoses in the data.

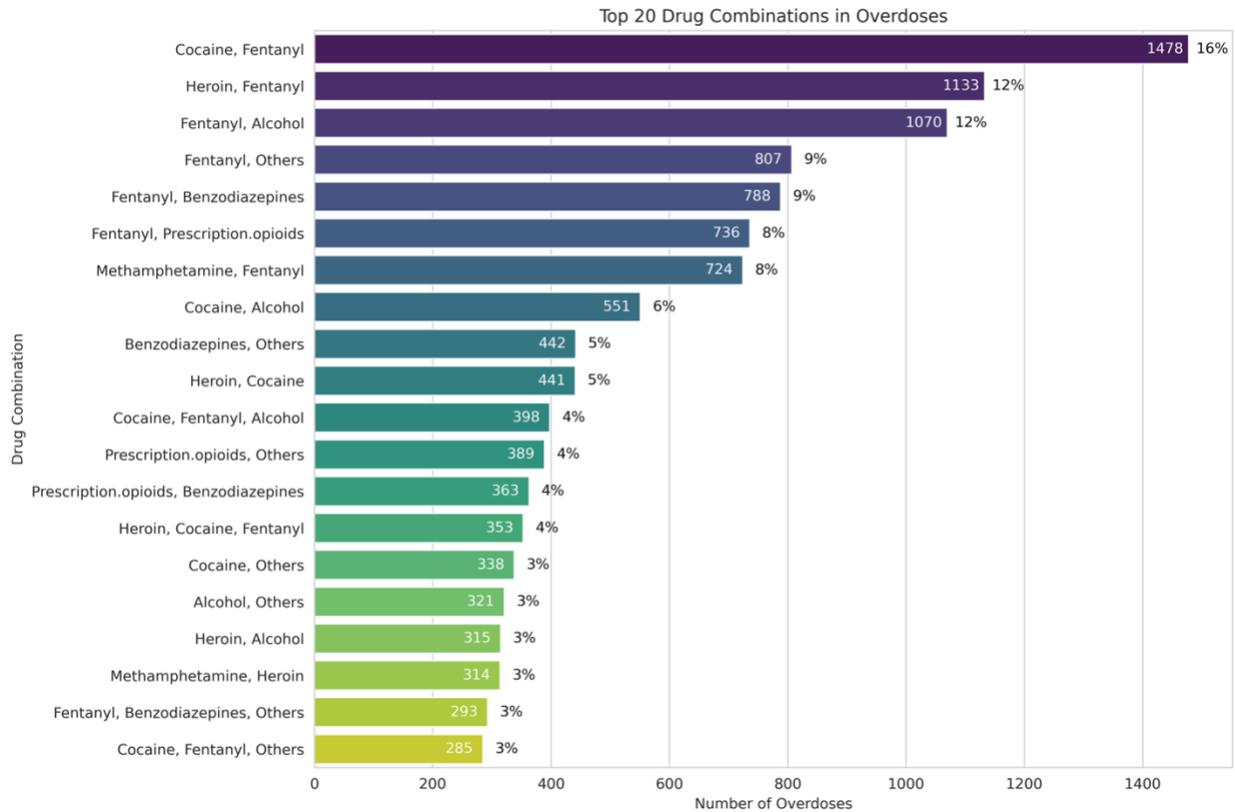

*FIGURE 1: The frequency of the different co-occurrences of drugs in the dataset showing that cocaine and fentanyl were the most common*

**Single Label Results**

The validation results were used to determine which models to evaluate on the test set. Results from the evaluation are seen in Supplemental Tables 1 and 2. Near-perfect performance was seen in classifying fentanyl and methamphetamine related deaths, achieving F1-scores of 0.997 (95% CI: 0.996-0.997) and 0.953 (95% CI: 0.952-0.955), respectively. Cocaine F1-score 0.976 (95% CI: 0.975-0.977), heroin F1-score 0.962 (95% CI: 0.961-0.963), and alcohol F1-score 0.934 (95% CI: 0.933-0.935) also showed high classification performance. By contrast, the model's performance was lower in the "Others" category, where the F1 score was 0.773 (95% CI: 0.769-0.776); prescription opioids also had a comparatively lower macro F1 Score of 0.840 (95% CI: 0.837-0.843).

## Multi Label Results

Table 2 shows a comparison between the single-label, multi-label, and fine-tuned BERT models on the test set. The single-label models combined using the BioclinicalBERT embeddings were the best performing of the single- and multi-label classic machine learning approaches, with a macro F1 of 0.929 (95% CI: 0.929-0.930) and an accuracy of 0.942 (95% CI: 0.941-0.942). The random forest and XGBoost multi-label models performed the worst with F1 Scores of 0.657 (95% CI: 0.656-0.657) and 0.790 (95% CI: 0.789-0.791), respectively. Overall, the fine-tuned BERT models performed best, with a macro F1 of ≥0.998 and an accuracy of 0.999 (95% CI: 0.998-0.999). The fine-tuned BioclinicalBERT model outperformed the finetuned BERT model marginally, scoring slightly higher in macro F1-score and accuracy with the BERT model achieving an F1 score of 0.992 (95% CI: 0.992-0.993) and the BioClinicalBERT model achieving a score of 0.998 (95% CI: 0.994-0.994).

Table 2: Comparison of classic single label, multi-label models and fine-tuned Bidirectional Encoder Representations from Transformers (BERT) models on the internal test set (best models are highlighted) *in the case of hamming loss a smaller value indicates a more performant model.

|  | Single label | | Multi label | | Fine tuned | |
|---|---|---|---|---|---|---|
| **Metric** | Model Combined (XGBoost, SVC) (CUIs) | Model Combined (XGBoost, SVC) (bcBERT embeddings) | Random Forest Multi Label (bcBERT embeddings) | XGBoost Multi Label (bcBERT embeddings) | Fine-tuned BERT | Fine-tuned bioclinicalBERT |
| *F1 Score* | 0.930 (0.930-0.931) | 0.929 (0.929-0.930) | 0.657 (0.656-0.657) | 0.790 (0.789-0.791) | 0.996 (0.996-0.996) | 0.998 (0.998-0.998) |
| *Accuracy* | 0.941 (0.940-0.941) | 0.942 (0.941-0.942) | 0.818 (0.817-0.818 | 0.892 (0.891-0.892) | 0.997 (0.997-0.997) | 0.998 (0.998-0.999) |
| *AUROC* | 0.982 (0.982-0.982) | 0.998 (0.998-0.998) | 0.980 (0.980-0.980) | 0.992 (0.991-0.992) | 1.00 (1.00-1.00) | 1.00 (1.00-1.00) |
| *Hamming Loss\** | 0.009 (0.009-.009) | 0.007 (0.007-0.007) | 0.055 (0.055-0.055) | 0.018 (0.018-0.018) | 0.00027 (0.00026-0.00027) | 0.00016 (0.00015-0.0017) |
| Abbreviations: AUROC, area under the receiver operating curve; BERT Bidirectional Encoder Representations from Transformers; bcBERT, BioClinicalBERT; CUI, concept unique identifier; XGBoost, Extreme Gradient Boosting; SVC, Support Vector Classifier | | | | | | |

### Other Large Language Models Results

Table 3 shows the results of the two best-performing large language models (LLMs) on the internal dataset. The Qwen-3 1.7B-Coroner (3-shot) model achieved a macro-average F1 score of 0.984 (95% CI: 0.980-0.986), while the Llama-3.2 3B-Instruct-Coroner (0-shot) model attained a slightly higher macro-average F1 score of 0.994 (95% CI: 0.992-0.996). While both models demonstrate strong internal performance, BioClinicalBERT outperformed both, with a near-perfect macro-average F1 score of 0.998 on the internal dataset (Table 3).

**Table 3: Performance metrics on the external test set of the finetuned BERT models**

| Model | F1 Score | Accuracy | Hamming loss* | AUROC |
| --- | --- | --- | --- | --- |
| *BioclinicalBERT* | 0.966 (0.966-0.967) | 0.964 (0.931-0.964) | 0.005 (0.005-0.005) | 0.994 (0.993-0.994) |
| BERT | 0.723 (0.721-0.725) | 0.884 (0.883-0.884) | 0.037 (0.037-0.038) | 0.951 (0.950-0.951) |
| * A lower hamming loss indicates a better model. Abbreviations: BERT, Bidirectional Encoder Representations from Transformers; AUROC, area under the receiver operating curve | | | | |

### External Validation

When evaluated on the external test set, the Qwen-3 1.7B-Coroner (3-shot) model outperformed all other LLMs, achieving a macro-average F1 score of 0.968 (95% CI: 0.961-0.974). In contrast, Llama-3.2 3B-Instruct-Coroner (0-shot) scored 0.959 (95% CI: 0.953-0.966). Qwen3 had the best performance on the external dataset marginally, scoring 0.002 higher than BioClinicalBERT on average which had a macro-average F1 score of 0.966 (95% CI: 0.966-0.967), as shown in Table 3. Notably, there was a substantial performance gap between BioClinicalBERT and the baseline BERT model on the external dataset (macro-average F1 scores of 0.966 vs. 0.723, respectively). This highlights the importance of both domain-specific pretraining and task-specific fine-tuning for effective generalization to unseen data.

**Table 4: Performance metrics of the two large language modles (LLMs) that performed best on the internal and external test sets**

| Model | Dataset | F1 Score | Accuracy |
|---|---|---|---|
| Qwen-3 1.7B Coroner (3 Shot) | External Test | 0.968 (0.961 - 0.974) | 0.958 (0.952 - 0.964) |
|  | Test | 0.984 (0.980 - 0.986) | 0.981 (0.977 - 0.985) |
| Llama-3.2 3B Instruct Coroner (0 shot) | External Test | 0.959 (0.953 - 0.966) | 0.948 (0.940 - 0.954 |
|  | Test | 0.994 (0.992 - 0.996) | 0.991 (0.988 - 0.994) |

**Error analysis**

**Internal Test Set**

The internal test set consisted of 7,054 cases; of these cases, the fine-tuned BioclinicalBERT (best-performing model) classified 11 cases incorrectly. The most frequent misclassification was the "Others" category (n=3), reflecting its broad nature, often triggered by the presence of multiple drugs, particularly benzodiazepines. Benzodiazepine errors (n=3) included misidentifying benzothiazepine and failing to detect flualprazolam. Other notable errors included false positives for alcohol (n=2), likely due to co-ingestion or general toxicity mentions. Supplemental Table 4 breaks down the number of mistakes by category.

**External Test Set**

In the external test set, the fine-tuned BioclinicalBERT model classified 116 cases incorrectly. Similarly, the most frequent misclassification was the same as the internal dataset, the "Others" category (n=84). This issue dominated the errors with 12 false positives and 72 false negatives; the model was oversensitive when many different substances occurred and was unable to detect some unfamiliar drug names. The next most problematic category was the "Any Drugs" category (n=24), the model missed generic mentions of drug use, especially when no specific drug is mentioned but drug use is implied e.g. "DRUG(S) TOXICITY". Benzodiazepines were the third most error prone category (n=11), where the model confused some benzothiazepines with benzodiazepines. Supplemental Table 5 breaks down the misclassifications by category and some possible reasons for these.

**Discussion**

In this study, we demonstrated that NLP models can accurately classify specific drug involvement from unstructured medical examiner text. Notably, the highest-performing model (BioClinicalBERT) achieved near-perfect macro-average F1 scores across both internal and external validation datasets, underscoring the importance of domain-specific pretraining and task-specific fine-tuning. These findings build on prior research by addressing key limitations of earlier approaches (4), such as reliance on concept unique identifiers (CUIs) or limited substance classification scope. Importantly, our results suggest that using contextualized embeddings derived from clinical data enables robust classification even in the presence of spelling variations, co-occurring substances, and complex phrasing common in medical examiner reports. This approach also offers substantial improvements in speed and scalability for public health surveillance: the model classified 7,088 cases in just 9.02 seconds - approximately 1.28 seconds per 1,000 cases, all while maintaining excellent accuracy. This suggests that such models could enable coroner offices to clear the backlog of cases and enable up to date reporting on overdose mortality; overcoming delays associated with manual ICD-10 coding and enhancing timely response to emerging drug trends.

Although contemporary decoder-only LLMs may offer improved performance in several NLP tasks due to their broader linguistic capabilities and exposure to more diverse training data (13,24,25), the LLMs tested in this paper did not show consistent or substantial improvement over earlier BERT-style models. In this work, we limited the LLMs used to a maximum of 8 billion parameters. This decision reflects a key consideration: the computational resources and costs associated with deploying LLMs may be prohibitive for medical examiner and coroner offices, many of which operate with limited technical infrastructure (26). In contrast, BERT-style encoder models strike a pragmatic balance between performance and feasibility. These models are naturally suited for classification tasks, require less computational power, and when fine-tuned on domain-specific data achieve near state-of-the-art performance. Given the well-defined nature of overdose classification and the strong results demonstrated here, our findings suggest that smaller, encoder-based models offer an efficient and scalable solution for accelerating overdose surveillance efforts across jurisdictions.

**Limitations**

Despite these promising results, several limitations warrant consideration. First, although our external validation dataset included jurisdictions not represented in the training data, the overall geographic scope was still limited to selected U.S. counties and may not capture variation in language use across all regions or among different coronial systems (27). Second, while the model performed well across most drug categories, performance was lower for substances grouped into the "Other" category, which likely reflects the heterogeneity and rarity of individual drugs within that class. Third, although we excluded chronic disease-related deaths, there remains some subjectivity in manual annotation of cause-specific drug involvement, which could introduce bias into the reference standard. Lastly, this study evaluated retrospective data; future work is needed to assess how these models perform in real-time operational settings, particularly when novel substances emerge with unfamiliar terminology.

**Conclusion**

We demonstrated that NLP models, especially their transformer-based domain-specific variants (BioClinicalBERT), were highly effective in identifying overdose death cases from textual data. Our results show near-perfect accuracy (99.8%) in classifying overdose cases from textual data and a robust F1 score on an external dataset (96.6%). This performance supports their use in automating overdose classification workflows, enabling faster and more scalable public health response systems. These findings contribute to a growing body of work demonstrating the effectiveness of NLP in classifying medical data, and in this case, medical examiner data. By enabling high-accuracy, real-time classification, these models can support public health authorities in detecting new substance trends and facilitating timely, targeted interventions in overdose crises.

# SUPPLEMENTAL MATERIALS

### Explainability: Methods

Explaining the final model was performed using the transformers-interpret library. This was used to verify the detection of drug-related linguistic patterns and enhance model interpretability. This library utilizes integrated gradients, attributing importance scores to input tokens by measuring their impact on model predictions through gradient-based analysis (28). These gradients are shown as highlights of green or red on tokens. The intensity of this highlighting relates to how strongly the given token contributed to the final prediction, red indicating a negative contribution to the probability and green a positive contribution.

### Explainability: Results

The following error analysis was carried out using the transformer-interpret (28) library as detailed in our Methods. The "#" indicates when a token is part of a word and has been broken up. This enables to see which parts of the words are important and reveals patterns detected in the linguistic structure of the drugs.

Two examples of mislabeled data that the model picked up on and correctly classified are shown in Supplemental Figure 1 and 2. The example in Figure 1 highlights a spelling mistake of the benzodiazepine name "flualprazolam." The true label for benzodiazepine, in this case, was 0, but the predicted probability was 0.58 for the benzodiazepine class. Therefore, the model still correctly classified this as an overdose relating to benzodiazepine ingestion.

Supplemental Figure 2 shows that the tokens 'sequel' and 'injury' contributed negatively to the final probability, leading to a prediction of 0 rather than 1. This input is labeled as an overdose via prescription opioids. The label is incorrect; reading the input: the oxycontin use was historical, and the person died through brain injury; this is also suggested by the "history, drug, intake" parts of the input. The model was able to see this where the labeler was not.

**SUPPLEMENTAL MATERIALS FIGURES**

acute f#ent#any#l flu#ap#raz#ola#m and cocaine toxicity ing#est#ion of drugs

**SUPPLEMENTAL FIGURE 1:** A visualization of the model's decision process which led to the prediction of a benzodiazepine related overdose

sequel#ae of an an#oc#ic brain injury o#xy#con#tin intake by history

**SUPPLEMENTAL FIGURE 2:** A visualization of the model's decision process which led to the final prediction to conclude this was death by brain injury and not overdose.

**Supplemental Table 1:** Metrics describing the performance of the single label classifiers, comparison of different embedding methods when paired with different model architectures.

| Substance | Model | BioClinicalBERT embeddings F-Score | GloVe Embeddings F-Score | CUI2Vec Embeddings F-Score |
|---|---|---|---|---|
| Any Opioid | SVM | 0.986 | 0.989 | 0.992 |
| | XGBoost | 0.963 | 0.983 | 0.990 |
| | Random Forest | 0.949 | 0.970 | 0.986 |
| | Logistic Regression | 0.984 | 0.988 | 0.991 |
| Heroin | SVM | 0.984 | 0.988 | 1.000 |
| | XGBoost | 0.884 | 0.960 | 0.988 |
| | Random Forest | 0.773 | 0.896 | 0.962 |
| | Logistic Regression | 0.957 | 0.976 | 0.993 |
| Fentanyl | SVM | 0.998 | 0.999 | 0.999 |
| | XGBoost | 0.965 | 0.989 | 0.998 |
| | Random Forest | 0.942 | 0.973 | 0.997 |
| | Logistic Regression | 0.993 | 0.997 | 0.998 |
| Prescription Opioid | SVM | 0.916 | 0.944 | 0.991 |
| | XGBoost | 0.782 | 0.904 | 0.950 |
| | Random Forest | 0.679 | 0.808 | 0.896 |
| | Logistic Regression | 0.868 | 0.913 | 0.961 |
| Methamphetamine | SVM | 0.980 | 0.981 | 0.998 |
| | XGBoost | 0.795 | 0.945 | 0.959 |
| | Random Forest | 0.809 | 0.929 | 0.963 |
| | Logistic Regression | 0.972 | 0.975 | 0.976 |
| Cocaine | SVM | 0.992 | 0.987 | 0.995 |
| | XGBoost | 0.818 | 0.954 | 0.982 |
| | Random Forest | 0.807 | 0.890 | 0.955 |

|  | Logistic Regression | 0.981 | 0.973 | 0.987 |
| --- | --- | --- | --- | --- |
| Benzodiazepine | SVM | 0.949 | 0.752 | 0.752 |
|  | XGBoost | 0.790 | 0.862 | 0.679 |
|  | Random Forest | 0.697 | 0.733 | 0.672 |
|  | Logistic Regression | 0.895 | 0.723 | 0.743 |
| Alcohol | SVM | 0.968 | 0.945 | 0.915 |
|  | XGBoost | 0.885 | 0.954 | 0.914 |
|  | Random Forest | 0.760 | 0.949 | 0.917 |
|  | Logistic Regression | 0.962 | 0.939 | 0.914 |
| Others | SVM | 0.843 | 0.767 | 0.805 |
|  | XGBoost | 0.714 | 0.834 | 0.763 |
|  | Random Forest | 0.721 | 0.760 | 0.775 |
|  | Logistic Regression | 0.841 | 0.755 | 0.776 |

Abbreviations: BERT, Bidirectional Encoder Representations from Transformers; CUI, Concept Unique Identifiers; GloVe, Global Word Embedding Vectors; SVM, Support Vector Machine; XGBoost, Extreme Gradient Boosting

**Supplemental Table 2: Test Results Single Label Means with 95% Confidence Intervals (best model found on validation set: Support Vector Machine (SVM))**

| Metric | Mean (95% Confidence Interval) | | | | | | | | |
| --- | --- | --- | --- | --- | --- | --- | --- | --- | --- |
| | Any opioid | Heroin | Fentanyl | Prescription Opioids | Meth | Cocaine | Benzodiazepines | Alcohol | Others |
| F-Score | 0.982 (0.982-0.983) | 0.962 (0.961-0.963) | 0.997 (0.996-0.997) | 0.840 (0.837-0.843) | 0.953 (0.952-0.955) | 0.976 (0.975-0.977) | 0.892 (0.890-0.895) | 0.934 (0.933-0.935) | 0.773 (0.769-0.776) |
| AUROC | 0.999 (0.999-1.00) | 0.998 (0.997-0.998) | 1.000 (1.000-1.000) | 0.996 (0.996-0.996) | 1.000 (1.000-1.000) | 0.998 (0.998-0.998) | 0.997 (0.997-0.997) | 0.999 (0.999-0.999) | 0.992 (0.992-0.993) |
| AUPRC | 0.998 (0.998-0.998) | 0.984 (0.983-0.985) | 1.000 (1.000-1.000) | 0.942 (0.940-0.944) | 0.995 (0.994-0.995) | 0.992 (0.991-0.992) | 0.958 (0.956-0.960) | 0.992 (0.991-0.992) | 0.875 (0.872-0.879) |

Abbreviations: AUROC, area under the receiver operator curve; AUPRC, area under the precision recall curve

**Supplemental Table 3: Full list of large language models (LLMs) tested and their macro-average F1 scores**

| Model | Finetuned | 0-shot | 3-shot | 5-shot | 10-shot |
|---|---|---|---|---|---|
| Llama-3.2-1B-Instruct | No | 0.261 | 0.205 | 0.242 | 0.277 |
| Llama-3.2-1B-Instruct | Yes | 0.951 | 0.739 | 0.645 | 0.661 |
| Llama-3.2-3B-Instruct | No | 0.819 | 0.849 | 0.806 | 0.808 |
| Llama-3.2-3B-Instruct | Yes | 0.959 | 0.844 | 0.787 | 0.757 |
| Llama-3.1-8B-Instruct | No | 0.924 | 0.888 | 0.883 | 0.878 |
| Llama-3.1-8B-Instruct | Yes | 0.959 | 0.932 | 0.926 | 0.921 |
| Qwen3-0.6B | No | 0.447 | 0.469 | 0.498 | 0.514 |
| Qwen3-0.6B | Yes | 0.916 | 0.899 | 0.871 | 0.866 |
| Qwen3-1.7B | No | 0.731 | 0.79 | 0.795 | 0.796 |
| Qwen3-1.7B | Yes | 0.959 | 0.968 | 0.965 | 0.964 |
| Qwen3-4B | No | 0.938 | 0.92 | 0.928 | 0.886 |
| Qwen3-4B | Yes | 0.962 | 0.961 | 0.958 | 0.963 |
| Qwen3-8B | No | 0.944 | 0.936 | 0.939 | 0.942 |
| Qwen3-8B | Yes | 0.946 | 0.939 | 0.943 | 0.942 |

**Supplemental Table 4: Error analysis and reasoning on internal dataset**

| Drug Class | FP | FN | Total Errors | Possible Reasons |
|---|---|---|---|---|
| Methamphetamine | 1 | 0 | 1 | Occasionally detects MDMA as methamphetamine due to shared naming |
| Heroin | 1 | 0 | 1 | Associating other drugs with combined heroin use as well as cardiovascular related wording |
| Cocaine | 0 | 0 | 0 | N/A |
| Fentanyl | 0 | 0 | 0 | N/A |
| Alcohol | 2 | 0 | 2 | Associating "toxicity" or the combination of many drugs with alcohol |
| Prescription Opioids | 0 | 0 | 0 | N/A |
| Any Opioids | 0 | 1 | 1 | Unable to detect "ISOTONITAZENE" as an opioid |
| Benzodiazepines | 2 | 1 | 3 | Detecting a benzothiazepine as a benzodiazepine, unable to detect some benzos |
| Others | 3 | 0 | 3 | Overly sensitive especially when many different drugs are mentioned, mostly this occurs when there are lots of benzos also present |
| Any Drugs | 0 | 0 | 0 | N/A |
| **Total** | **9** | **2** | **11** | N/A |

**Supplemental Table 5: Error analysis and reasoning on external dataset**

| Drug Class | FP | FN | Total Errors | Possible Reasons |
|---|---|---|---|---|
| Methamphetamine | 6 | 2 | 8 | Detecting MDMA as methamphetamine when written in full |
| Heroin | 4 | 0 | 4 | Confusion arises when many different substances mentioned, heroin is predicted although not in the substances |
| Cocaine | 4 | 2 | 6 | Associates drug use and cardiovascular use with cocaine rarely classifies this as cocaine use |
| Fentanyl | 0 | 0 | 0 | N/A |
| Alcohol | 0 | 6 | 6 | Misses mentions of cocaethylene, where alcohol or ethanol is not specifically mentioned |
| Prescription opioids | 0 | 0 | 0 | N/A |
| Any Opioids | 0 | 3 | 3 | Missed mentions of mitragynine as an opioid |
| Benzodiazepines | 4 | 7 | 11 | Misclassified benzothiazepines as benzodiazepines and missed some generic mention of benzodiazepines |
| Others | 12 | 72 | 84 | Overly sensitive when many different drugs are present, in the false negative case the model misses acetaminophen and some antihistamines |
| Any Drugs | 3 | 21 | 24 | Mention of drug use without specific mention of which drugs causes the model to miss these cases |
| Total | 33 | 113 | 146 | N/A |